\documentclass[sigconf,nonacm]{aamas}

\usepackage{algorithm}
\usepackage{algorithmic}
\usepackage{booktabs}
\usepackage{amsmath}
\usepackage{xcolor}
\usepackage{enumitem}

\newcommand{\CaptionWithDescription}[1]{%
  \caption{#1}%
  \Description{#1}%
}

\usepackage{newfloat}
\usepackage{listings}
\DeclareCaptionStyle{ruled}{labelfont=normalfont,labelsep=colon,strut=off} % DO NOT CHANGE THIS
\floatstyle{ruled}
\newfloat{listing}{tb}{lst}{}
\floatname{listing}{Listing}

\usepackage{balance} % for balancing columns on the final page

\setcopyright{ifaamas}
\acmConference[AAMAS '26]{Proc.\@ of the 25th International Conference
on Autonomous Agents and Multiagent Systems (AAMAS 2026)}{May 25 -- 29, 2026}
{Paphos, Cyprus}{C.~Amato, L.~Dennis, V.~Mascardi, J.~Thangarajah (eds.)}
\copyrightyear{2026}
\acmYear{2026}
\acmDOI{}
\acmPrice{}
\acmISBN{}


\title[Break Out the Silverware: Semantic Understanding of Stored Household Items]{%
  \texorpdfstring{Break Out the Silverware:\\Semantic Understanding of Stored Household Items}%
  {Break Out the Silverware: Semantic Understanding of Stored Household Items}%
}

\author{Michaela Levi Richter}
\affiliation{
  \institution{Bar Ilan University, Computer Science Department}
  \city{Ramat Gan}
  \country{Israel}}
\email{michaela.levi20@gmail.com}

\author{Reuth Mirsky}
\affiliation{
  \institution{Tufts University, Department of Computer Science}
  \city{Medford, MA}
  \country{USA}}
\email{reuth.mirsky@tufts.edu}

\author{Oren Glickman}
\affiliation{
  \institution{Bar Ilan University, Computer Science Department}
  \city{Ramat Gan}
  \country{Israel}}
\email{oren.glickman@biu.ac.il}

\begin{abstract} 
``Bring me a plate.'' For domestic service robots, this simple command reveals a complex challenge: inferring where everyday items are stored, often out of sight in drawers, cabinets, or closets. Despite advances in vision and manipulation, robots still lack the commonsense reasoning needed to complete this task. We introduce the \textbf{Stored Household Item Challenge}, a benchmark task for evaluating service robots’ cognitive capabilities: given a household scene and a queried item, predict its most likely storage location.
Our benchmark includes two datasets: (1) a real-world evaluation set of 100 item-image pairs with human-annotated ground truth from participants' kitchens, and (2) a development set of 6,500 item-image pairs annotated with storage polygons over public kitchen images. These datasets support realistic modeling of household organization and enable comparative evaluation across agent architectures.
To begin tackling this challenge, we introduce \textbf{NOAM (Non-visible Object Allocation Model)}, a hybrid agent pipeline that combines structured scene understanding with large language model inference. NOAM converts visual input into natural language descriptions of spatial context and visible containers, then prompts a language model (e.g., GPT-4) to infer the most likely hidden storage location. This integrated vision-language agent exhibits emergent commonsense reasoning and is designed for modular deployment within broader robotic systems.
We evaluate NOAM against baselines including random selection, vision-language pipelines (Grounding-DINO + SAM), leading multimodal models (e.g., Gemini, GPT-4o, Kosmos-2, LLaMA, Qwen), and human performance. NOAM significantly improves prediction accuracy and approaches human-level results, highlighting best practices for deploying cognitively capable agents in domestic environments. \end{abstract}


\newcommand{\BibTeX}{\rm B\kern-.05em{\sc i\kern-.025em b}\kern-.08em\TeX}

\begin{document}

%%% The following commands remove the headers in your paper. For final 
%%% papers, these will be inserted during the pagination process.

\pagestyle{fancy}
\fancyhead{}

%%% The next command prints the information defined in the preamble.

\maketitle 

%%%%%%%%%%%%%%%%%%%%%%%%%%%%%%%%%%%%%%%%%%%%%%%%%%%%%%%%%%%%%%%%%%%%%%%%

\section{Introduction}
%Submission to CoRL 2025 will be entirely electronic, via a web site (not email). Information about the submission process and \LaTeX{} templates are available on the conference web site at \url{https://corl.org/}. For camera ready submission, use the \texttt{final} option for the \verb|\usepackage| command. 
% \begin{itemize}
%     \item Motivation: Service robots often operate in environments where household items are stored in drawers, cabinets, or closets, rather than being visibly accessible.
%     \item Problem Statement: Existing robotic perception systems do not effectively predict stored object locations.
%     \item Contribution: We propose a model that learns storage patterns to predict object locations in a new household.
% \end{itemize}

% Service robots are increasingly capable of navigating and manipulating objects in household environments. However, they still struggle with a core challenge: commonsense reasoning about where items are typically stored. For instance, when asked to fetch a mug, a human avoids searching inside the refrigerator or under the sink, instead relying on shared expectations. Translating this intuition into machine reasoning remains a major challenge.

Service robots are increasingly capable of navigating and manipulating objects in household environments. Yet, they remain limited in a fundamental way: commonsense reasoning about where items are typically stored when those items are not visible. For humans, such inference is effortless. When asked to fetch a mug, we do not search the refrigerator or under the sink; instead, we rely on deeply shared expectations about household organization. Translating this intuition into machine reasoning remains an unsolved problem.

We introduce the \textbf{Stored Household Item Challenge}, a new benchmark designed to evaluate a specific form of commonsense reasoning: predicting the likely storage location of household items that are not visible. Given an indoor scene and a queried item, the task is to infer which storage container (such as a drawer, cabinet, or closet) is most likely to contain the item, despite it being hidden from view. Unlike traditional object detection, which focuses on visible entities, this task requires reasoning about concealed items based on indirect visual cues and prior semantic knowledge. We hypothesize that household storage follows consistent patterns (for example, mugs are usually stored in upper cabinets rather than drawers) that can be learned and generalized.

% Current benchmarks overlook this type of reasoning. Vision-language models (VLMs) and object detectors perform well on visible items but struggle with inferring what is not seen. To our knowledge, there is no prior structured benchmark for fully hidden object localization. Existing datasets lack labels for expected-but-not-visible items, and VLMs are not trained to reason about functional storage norms or household organization. This challenge provides a testbed for studying such practical yet underexplored capabilities, especially relevant for robots in cluttered or unfamiliar spaces. 

\begin{figure}[t]
\vspace{4mm}
    \centering
    \includegraphics[clip, trim={0 100 0 0}, width=0.4\textwidth]{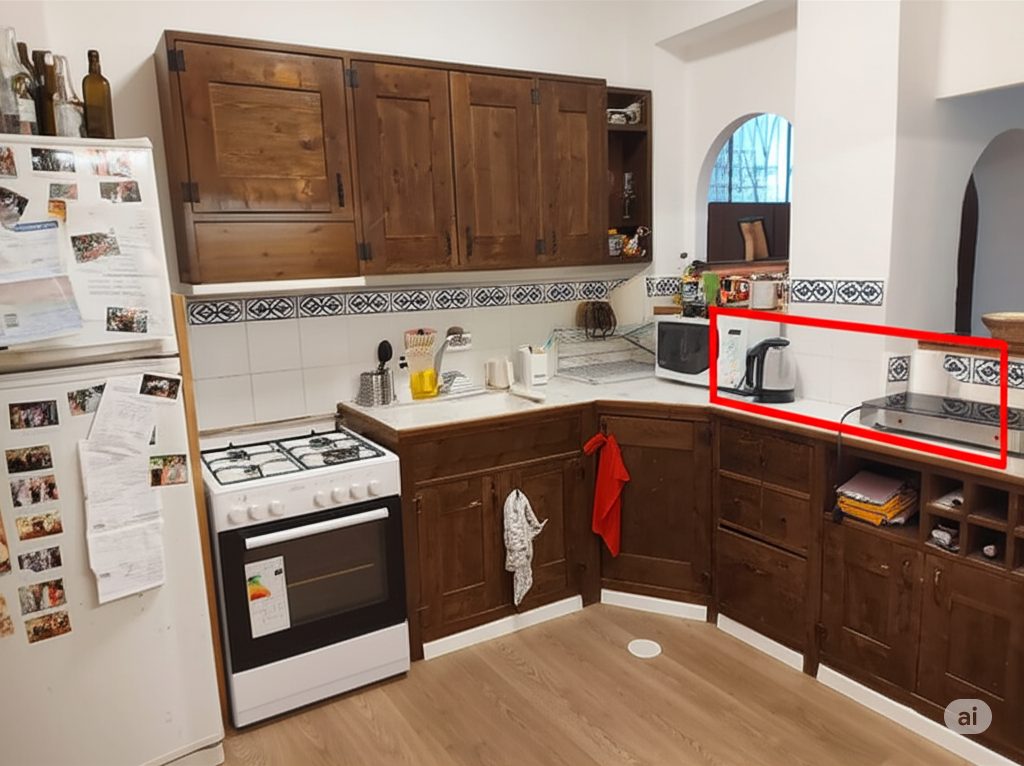}
    \CaptionWithDescription{Kitchen scene shown to Gemini, with its predicted spoon storage location highlighted in the bounding box.}
    \label{fig:gemini-spoon}
\end{figure}

\begin{figure}[t]
    \centering
    \includegraphics[clip, trim={0 100 0 0}, width=0.4\textwidth]{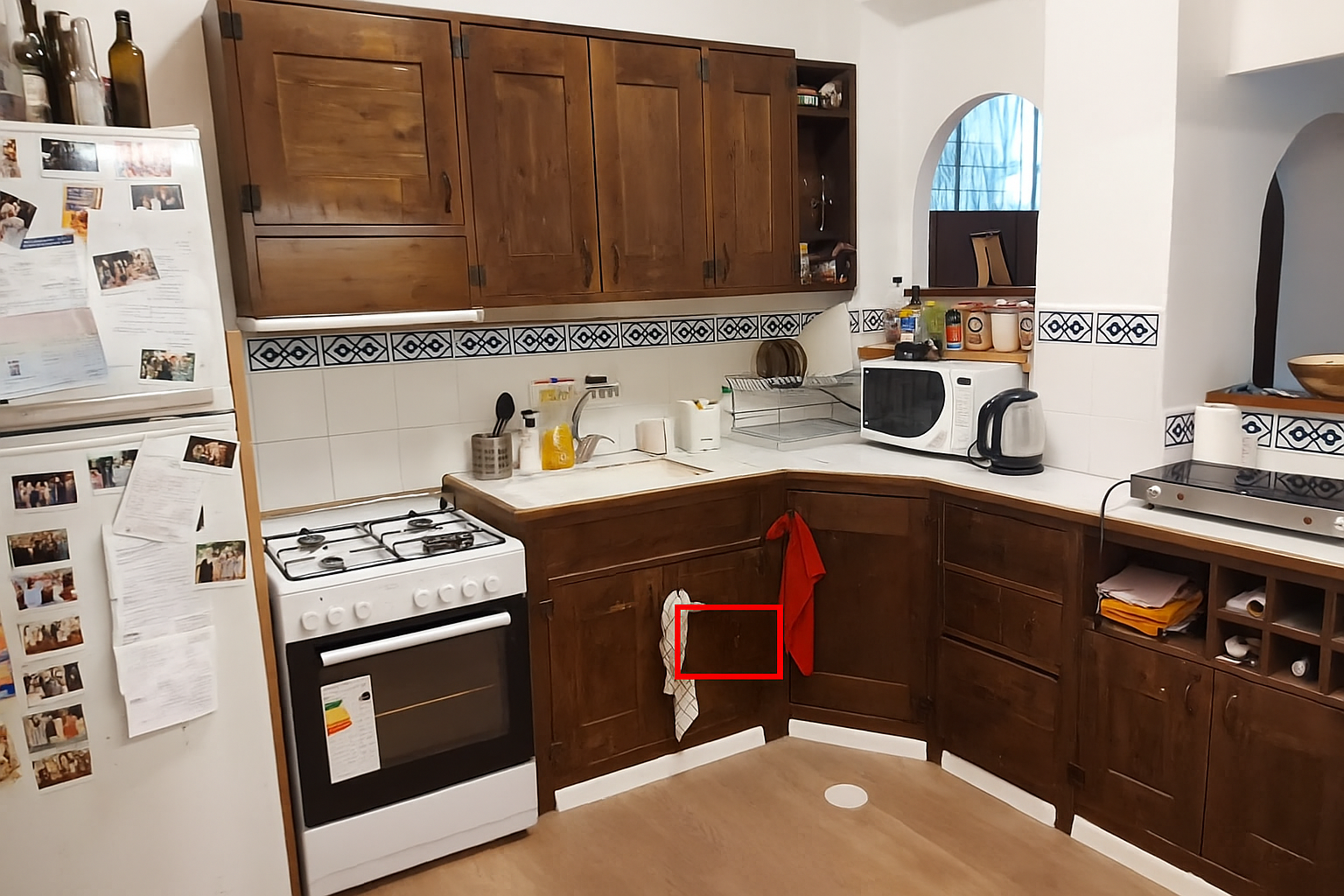}
    \CaptionWithDescription{Kitchen scene shown to GPT, with its predicted spoon storage location highlighted in the bounding box.}
    
    \label{fig:gpt-spoon}
\end{figure}

Current benchmarks overlook this reasoning dimension. Vision-Language Models (VLMs) and object detectors perform well on visible items but fail when asked to infer the invisible. For instance, we asked Gemini and GPT to identify where a spoon is most likely stored, using the following prompt:

\begin{quote}
\texttt{In the given image, which drawer or cabinet is the most likely place to store a spoon? Return the image with the exact drawer or cabinet clearly marked. Do not modify any other part of the image.}
\end{quote}

As shown in Figures~\ref{fig:gemini-spoon} and~\ref{fig:gpt-spoon}, both models struggled to select a plausible container. To our knowledge, no prior structured benchmark exists for this ability. Existing datasets lack labels for expected-but-not-visible items, and current models are not trained to reason about functional storage norms or household organization. This gap highlights both a scientific challenge and a practical barrier for embodied agents operating in real homes.

\begin{figure*}[t]
    \centering
    \includegraphics[trim=0 180 0 0, clip, width=\textwidth]{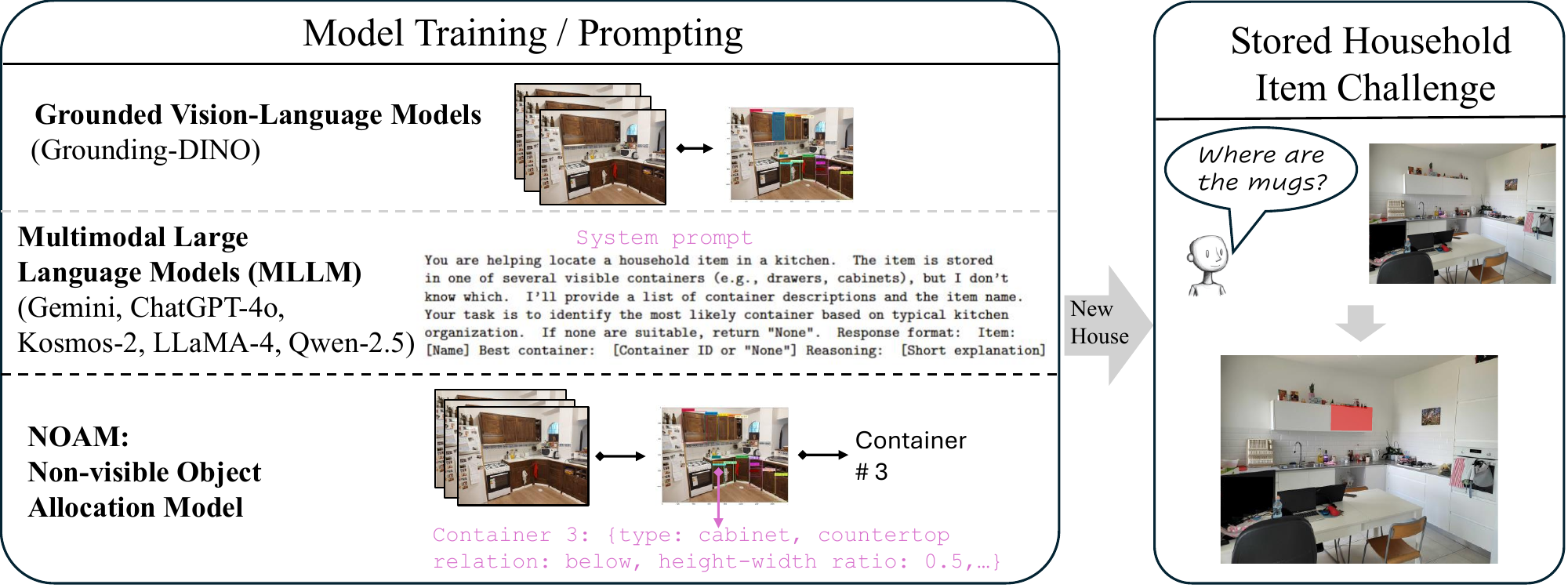}
    \CaptionWithDescription{Depiction of the Stored Household Item Challenge (right). We compare the performance of various models on this task (left), including grounded vision-language models, multimodal LLMs, and our model, NOAM.}
    \label{fig:intro}
\end{figure*}

To support this task, we construct two datasets: (1) a 6,500-pair development set based on real kitchen scenes (SUN dataset), with container annotations by humans; and (2) a 100-example evaluation set collected from real homes. These datasets enable scalable training and robust testing across varied environments. We benchmark random baselines, VLMs, Multimodal Large Language Models (MLLMs), and human predictions (Figure~\ref{fig:intro}, top and middle left). 

% We also introduce \textbf{NOAM} (Non-visible Object Allocation Model), a novel language-based pipeline that reframes the task as structured text-based reasoning (Figure~\ref{fig:intro}, bottom left). By describing scenes and potential storage candidates in natural language and querying LLMs, NOAM significantly outperforms vision-language models, though still below human performance. This result underscores the potential of language-driven inference for commonsense reasoning.

We also introduce \textbf{NOAM} (Non-visible Object Allocation Model), a novel language-based pipeline that reframes the task as structured text-based reasoning (Figure~\ref{fig:intro}, bottom left). By describing scenes and candidate storage options in natural language and querying LLMs, NOAM significantly outperforms VLMs, though still below human performance. This suggests that language-driven inference offers a promising path for equipping agents with commonsense reasoning skills.

% Rather than optimizing for marginal gains, our goal is to formalize a reasoning skill that current models lack: context-aware inference grounded in priors, affordances, and scene semantics. The \textit{Stored Household Item Challenge} captures a class of reasoning - predicting what is likely but unseen - that is essential for embodied agents yet absent from most benchmarks. As LLMs advance, we expect performance to improve, but the core challenge remains a meaningful test of real-world understanding.

Rather than chasing marginal detection gains, our aim is to formalize a reasoning capability absent in today’s models: context-aware inference about what is unseen but likely. The \textit{Stored Household Item Challenge} captures this missing skill that is crucial for robots in cluttered, unfamiliar spaces. As LLMs evolve, we expect progress, but the challenge will remain a meaningful test of real-world understanding.

To summarize, the contributions of this work are threefold:

\begin{enumerate}[label=\arabic*.]
    \item New challenge: the Stored Household Item Challenge. 
    \item An annotation pipeline and dataset of item-image pairs grounded in real household scenes.
    \item NOAM (Non-visible Object Allocation Model), a language-based model that reframes visual inference as a structured textual reasoning task. We compare it against baselines, including random choice, VLM, MLLMs and humans.
\end{enumerate}

% Ultimately, this work aims to equip service robots with richer semantic understanding and contextual awareness, allowing them to operate more effectively in unfamiliar or dynamically changing home environments. By inferring the likely locations of hidden items using visual cues and commonsense knowledge, robots can make faster decisions, reduce reliance on user input, and improve task success in everyday scenarios. This ability is especially important for long-term autonomy, where robots must adapt to changing layouts and user habits over time. We invite the community to explore this new problem space, extend our framework, and help advance the development of more intelligent, adaptive household robots.

%===============================================================================

\section{Related Work}
% Our research focuses at the intersection of three main topics: Domestic Service Robots, Scene Understanding and Object Detection. As detailed below, in some parts our research relies on novel contributions, and in some parts our research will use existing methods. 
% Moreover, under object detection, we will use applications of vision and language models for semantic understanding, as detailed in this section.

% \begin{center}
%     \includegraphics[width=0.4\textwidth]{Venn of related work}
%     \captionof{figure}{Related work relevant topics}
% \end{center}

 % Adjust the width as needed
Our research lies at the intersection of three areas: domestic service robots, scene understanding, and object detection. While our work introduces a new approach to reasoning about hidden items, it builds on established methods across these domains, particularly recent advances in vision, vision-language, and MLLMs to enhance semantic understanding in complex domestic environments.

\paragraph{Domestic Service Robots}
Research on cleaning robots has expanded beyond vacuuming to include service robots capable of tidying, organizing groceries, washing dishes, and setting tables. Prior work focuses on learning processes tailored to specific homes \cite{hasegawa_2023_pl, kim_control_2019, shah_deep_2020} and on user-instructed behaviors \cite{Kaneda2024learningtorank, khan_towards_2022, matsushima_world_2022, ribeiro_charmie_2021, Wu_2023_tidyBot, wu2023integrating, yan_quantifiable_2021}. Some approaches use GNNs to learn user preferences from observations \cite{kapelyukh2021house} or exploit object hierarchies and collaborative filtering \cite{abdo_robot_2015}. However, most pick-and-place tasks involve visible surfaces like shelves or tables \cite{hu2023knolling, shiba_object_2023, xu_tidy_2023}, whereas we target items stored in enclosed containers.

\citet{liu_service_2022} proposed a knowledge-based framework for object search using ontologies and multi-domain knowledge. \citet{ramrakhya2024seeing} define a Semantic Placement (SP) task for predicting object placements, but focus on visible items. Our task generalizes to any household object and emphasizes hidden storage.

\citet{avidan_housekeep_2022} use commonsense reasoning to rearrange objects in simulation, but rely on partial observability and do not target concealed items. \citet{kurenkov_2023_object_search} assumes access to prior environment layouts when searching in containers. Our method aims to reduce robot deployment time by enabling inference without prior maps.
Additional work explores contextual semantics for continuous sweeping \cite{ahmadi2006multi} or for organizing items by user preferences \cite{adeleye_contextually_2023, avidan_housekeep_2022, Wu_2023_tidyBot, wu2023integrating, xu_tidy_2023}. Some methods adapt NLP to guide object arrangement \cite{hu2023knolling}, which helps integrate LLMs to generalize storage norms even without explicit user preferences.

While our method is currently designed and evaluated outside the context of physical robots, our broader vision is to support real-world domestic service robots that reason about storage in a human-aligned way. Although we have not yet tested our system on physical robots, this application remains a key motivation and long-term goal for our research.

%\paragraph{Scene Understanding}
Scene understanding involves analyzing visual scenes through both geometric and semantic context, along with inter-object relationships \cite{Naseer_2019}. This holistic understanding is essential for robots to collaborate effectively with humans \cite{AarthiSceneUnderstandingSurvey2017}, going beyond object classification to functional comprehension \cite{ye2016i}.

Subtasks such as scene classification help label scene components \cite{patel_survey_2020, tong_review_2017}, and assistive robotics research focuses on reducing computational overhead for intuitive interfaces \cite{bousquet-jette_fast_2017}.

Scene understanding also intersects with Simultaneous Localization and Mapping (SLAM), which enables robots to build maps of unknown environments while localizing themselves \cite{cadena_past_2016, chen_overview_2022, garg_semantics_2020, singandhupe_review_2019}. Incorporating semantics into SLAM allows robots to reason about object meaning and location together, improving navigational accuracy and object localization.

%\paragraph{Object Detection} 
Object detection identifies semantic instances in images or video \cite{zou_object_2023}. Early systems used handcrafted features, but deep learning now enables models to learn discriminative features through improved architectures, training, and optimization \cite{zhao_object_2019}.

Our task involves segmentation, including semantic segmentation \cite{minaee_image_2021} and instance segmentation \cite{hafiz_survey_2020}. We use Grounding-DINO for open-set object detection via text prompts \cite{liu_grounding_2023}, and Segment Anything Model (SAM) for segmentation \cite{kirillov2023segany}. Performance is evaluated using accuracy and intersection over union (IoU) metrics \cite{ayala_guidelines_2024}.
Concealed Object Detection (COD), which segments camouflaged objects \cite{fan_concealed_2022}, is not applicable here, as it assumes visibility. Similarly, occlusion detection focuses on partially visible objects in controlled settings \cite{saleh_occlusion_2021}, whereas we target fully hidden items. Work in areas like prisons or airports uses thermal cameras for detecting occluded objects \cite{kristoffersen_pedestrian_2016}, but these technologies are unsuitable for locating stored household items.

Recent advances in zero-shot object detection further leverage VLMs and MLLMs to detect novel object categories using natural language prompts \cite{madan2024revisiting}. 

In our experiments, we compare a range of such models as baselines for the Commonsense Storage Prediction task.
%\paragraph{Vision and Language Models} We evaluate a diverse set of models spanning vision-only, vision-language, and multimodal large language models.
These models are used either to detect and segment storage containers or to predict likely storage locations of household items. Our dual strategy compares structured, text-based reasoning (NOAM) against end-to-end multimodal inference, enabling a comprehensive analysis of commonsense storage prediction capabilities.

%In our experiments, we compare a range of models as baselines for the Commonsense Storage Prediction task. These include vision-only models, vision-language pipelines, and recent multimodal large language models (MLLMs). Some are used to detect and segment storage containers directly, while others attempt to infer likely storage locations given the image and item query. This diversity of approaches enables a comprehensive evaluation of the task and highlights the strengths and limitations of existing vision and language systems in reasoning about hidden objects.

%===============================================================================

\section{The Stored Household Item Challenge}

We introduce the \textbf{Stored Household Item Challenge}, a benchmark for evaluating \emph{Commonsense Storage Prediction} in domestic environments. 
This task requires reasoning about \emph{non-visible} items. A system is to infer plausible hidden locations using visual context, semantic priors, and typical storage conventions. For example, given the query ``mug'', the model should prefer an upper cabinet over a lower drawer, even if the mug is not visible.

\textbf{Input:} A single RGB image of a scene and the name of a common household item (e.g., ``mug'', ``cutting board'').

\textbf{Output:} The most likely storage container \emph{instance} (e.g., a specific drawer or cabinet) in the image where the item is stored. The prediction must be grounded in a 2D polygon corresponding to a visible container.

\noindent This task involves three commonsense reasoning challenges:

\begin{itemize}
    \item \textbf{Reasoning Beyond Visibility:} Most vision systems detect only visible objects. This task requires inferring hidden item locations using indirect cues and commonsense knowledge.
    
    \item \textbf{Semantic Understanding and Commonsense Priors:} Success depends on functional knowledge of household norms combined with visual context.
    
    \item \textbf{Grounded, Actionable Output:} Predictions must be specific and physically actionable. Rather than generic guesses like ``a cabinet,'' the model must identify a concrete container instance suitable for robotic interaction.
\end{itemize}

To date, no existing dataset targets reasoning about hidden storage in realistic household scenes, and VLMs often default to literal descriptions of what is visible. By offering a structured task, curated data, and measurable evaluation criteria, this challenge fills a critical gap and provides a testbed for evaluating commonsense visual reasoning in service robotics.

%===============================================================================

\section{Datasets and Data Collection}

To predict where household items are typically stored, we required a dataset capturing commonsense storage behavior. Specifically, where items are likely kept out of sight in containers. As no existing dataset supported this task, creating a new one posed two challenges. First, in terms of \textbf{privacy}, domestic images often contain personal content that is difficult to anonymize and share ethically. Second, it is hard to \textbf{scale}. Gathering a large and diverse set of labeled images from volunteers is logistically challenging, making small-scale collection feasible but broad-scale coverage impractical. To address these challenges, we adopted a dual-dataset strategy:
\begin{itemize}
    \item A \textbf{crowdsourced development dataset} using public kitchen images, supporting design, tuning, and analysis.
    \item A smaller \textbf{real-world evaluation dataset} collected from participants’ homes, used for final testing.
\end{itemize}

The development set was built from the SUN dataset \cite{Xiao_SUN_2010, Xiao_SUN_2016}. Annotators recruited via the Upwork freelance platform\footnote{\url{https://www.upwork.com/}} selected the most likely storage container for a given item in each kitchen image. Figure \ref{fig:annotate} shows the annotation interface. We cleaned the data, removed duplicates, assessed annotator agreement, and curated 6,500 high-quality item-image pairs with polygon-level annotations.

%The evaluation set includes 100 item-image pairs from real kitchens. Each pair was labeled with the true storage location of a queried item and serves as ground truth.
The evaluation set comprises 100 item–image pairs collected from real kitchens, each annotated with the actual storage location of the queried item, serving as the ground-truth reference for model assessment.

To constrain variability, we focused on kitchens and selected 15 household items spanning common and uncommon examples: \textit{bottle opener, Tupperware containers, dish towels, cutting board, bowl, spices, spoon, mug, plate, pot, pan, cutting knife, cooking oil, screwdrivers}, and \textit{painkillers}. The full annotation protocol is released with this paper.

An IRB from \textit{redacted for anonymity} approved data collection involving human participants, covering both datasets.

\subsection{Dataset Preprocessing}

We applied Grounding-DINO \cite{liu_grounding_2023} and SAM \cite{kirillov_segment_2023} to segment all visible storage containers. Each development image contained $\sim$16 containers on average; each evaluation image, $\sim$19. These segmentations provide visual cues for detection and support automated answer verification, avoiding subjective free-form judgments.

\subsection{Development Dataset: Crowdsourced Annotations}
\label{sec:dev}

\begin{figure}[t]
    \centering
    \includegraphics[trim=0.5cm 0cm 1cm 0.5cm, clip, width=0.8\linewidth]{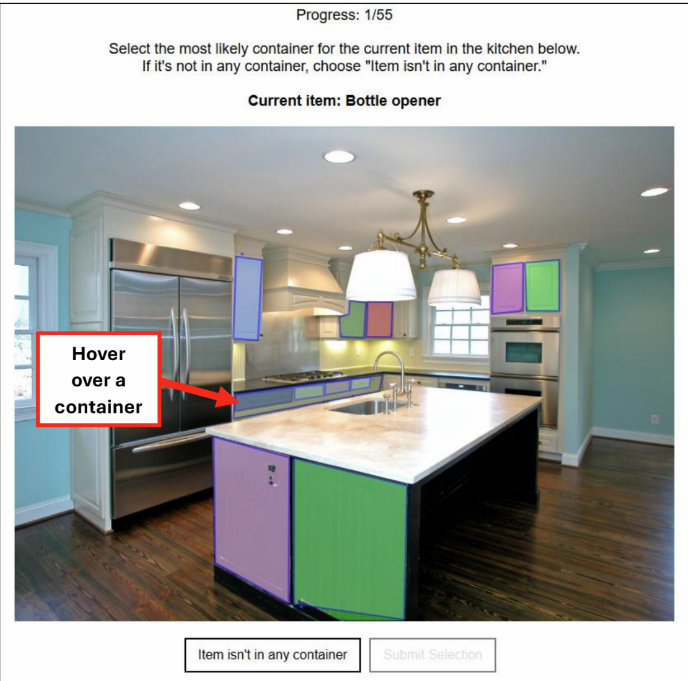}
    \CaptionWithDescription{A screenshot from the annotation tool used to collect human-labeled data efficiently.}
    \label{fig:annotate}
\end{figure}

From 1,746 kitchen images in SUN, we selected 1,656 with at least three detectable containers. For each of 13 items (excluding two held out for generalization tests), we sampled 500 images, totaling 6,500 item-image pairs. 

Three annotators (two from the U.S., one from Ireland) used a custom web tool to label images. Each image showed an item name and marked containers, as shown in Figure~\ref{fig:annotate}. Responses were stored in Firebase and local backups. Each annotator contributed 2,860 annotations (1,040 overlapping). Annotations were done in batches and manually reviewed. We collected 8,580 total annotations, from which the 6,500 development pairs were drawn.
To assess quality, 16\% of examples per item were triple-labeled. Agreement was computed using Fleiss' Kappa, detailed in Section~\ref{sec:clean}.

\subsection{Evaluation Data: Real-World Objects}
\label{sec:real}

We collected 100 item-image examples from 74 participant kitchens. Participants labeled storage locations for \textit{bottle openers, Tupperware containers, painkillers}, and \textit{screwdrivers}. The last two were excluded from the development set to test generalization. In addition, some participants labeled an extra item of their choice, beyond the four predefined categories. Participants were instructed to exclude sensitive content, and all images were manually reviewed for identifying details.
This dataset was used both to validate annotator reliability and as ground truth for model evaluation.

\subsection{Data Cleaning and Processing}
\label{sec:clean}

We applied the following procedures to ensure data quality:
\begin{enumerate}[label=\arabic*.]
    %\item Filtered submissions from unregistered IPs.
    \item Removed duplicates and conflicting responses.
    \item Consolidated multiple annotations:
    \begin{itemize}
        \item Majority vote was used when two annotators agreed.
        \item Otherwise, one label was selected at random.
    \end{itemize}
    \item Split data into development and evaluation sets.
\end{enumerate}

Inter-annotator agreement varied by item. Fleiss' Kappa was $\kappa = 0.354$ (fair) for the development set and $\kappa = 0.372$ for the evaluation set. On the development set, items with standardized locations (e.g., bottle openers, $\kappa = 0.494$; plates, $\kappa = 0.478$) showed higher agreement. Subjective categories like Tupperware had lower agreement ($\kappa = 0.27155$). These results highlight both the inherent difficulty and variability of the task, while also suggesting that people rely on more than random reasoning when searching for hidden household items.

%===============================================================================

\section{NOAM: Non-visible Object Allocation Model}
% \begin{itemize}
%     \item \textbf{End-to-End Learning Approach:} Using a vision-language model to predict storage locations.
%     \item \textbf{Composite Model:} Detecting storage containers first, then predicting item placement.
%     \item \textbf{Training Data and Labeling:} Dataset of household objects and common storage patterns.
%     \item \textbf{Key Challenges:} Handling noise, diverse household layouts, and domain adaptation.
% \end{itemize}

Our goal is to enable VLMs, and later on service robots, to reason about storage locations with accuracy approaching or exceeding human-level performance as measured on our dataset ($\sim$$30\%$). To this end, we reformulate the task from an image-text multimodal problem into a purely language-based one, casting it as a natural language understanding challenge.

\begin{figure*}
    \centering
    \includegraphics[trim=0 190 200 0, clip, width=\linewidth]{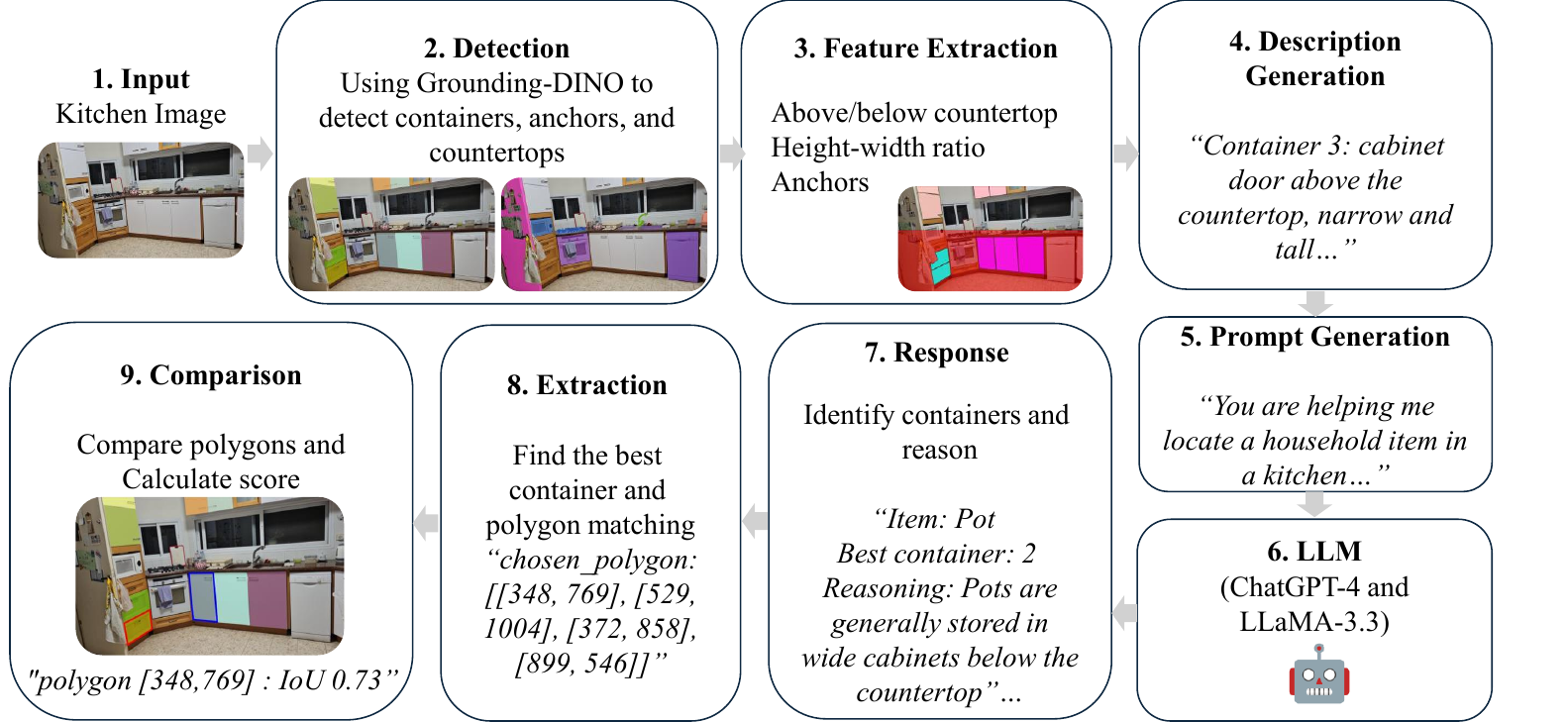}
    \CaptionWithDescription{An overview of our NOAM pipeline: from an input image (1) through container detection (2), feature extraction (3), textual explanation (4), prompt generation (5), through LLM processing (6) to response (7) and value extraction (8) and overall evaluation (9).}
    \label{fig:noam}
\end{figure*}

%Our goal is to locate stored items in kitchen environments with human-level accuracy ($\sim$$30\%$). To tackle this, we reformulate the task from image-text multimodal to a purely language-based problem, casting it as a natural language understanding challenge.
%
Our approach, the Non-visible Object Allocation Model (NOAM), is outlined in Figure~\ref{fig:noam}. For each container detected in the scene, we extract a set of visual and spatial features and translate them into clear, natural language descriptions (steps 2–3). These descriptions are incorporated into carefully structured prompts that instruct a language model (e.g., ChatGPT) to infer the most probable storage location for a given item (steps 4–5). The model then reasons to select the most probable container(s) (step 7). We extract the predicted container ID, defaulting to the first option when multiple candidates are returned, and map it to the corresponding polygon in the image (step 8). Finally, spatial accuracy is evaluated using Intersection over Union (IoU) with the ground truth (step 9). 
%Full results are presented in Section~\ref{sec:result}.

\paragraph{Feature Extraction}

To represent each container in an image, we extracted features relevant to reasoning, including:
container type (e.g., drawer, cabinet door) and confidence score, position relative to the countertop (above/below), neighboring containers and anchor objects (e.g., sink, oven), closest container to each anchor, aspect ratio and shape, and other contextual cues.

Some features were computed directly from container polygons, while others required re-applying Grounding-DINO and SAM to detect countertops and anchor objects. Each container was then represented as a row in a \textit{container information table}, initially containing image path and polygon data.
We augmented the table with the extracted features through the following steps:
\begin{enumerate}[label=\arabic*.]
    \item Assign global and local container IDs.
    \item Extract and assign labels and confidence scores.
    \item Resolve ambiguous labels (e.g., ``drawer cabinet door'') using context from neighboring containers.
    \item Compute container dimensions and aspect ratio.
    \item Determine spatial relation to the countertop.
    \item Identify neighboring containers and anchor objects.
    \item Measure distances and angles from each container to all anchors to determine the closest container to each anchor.
\end{enumerate}

\paragraph{Natural Language Description}

This stage involves two main steps: first, selecting the most informative and relevant features to include in each container’s description, with a focus on attributes that help the language model infer spatial and categorical relationships; and second, converting these features into clear and intuitive natural language. Rather than using raw numerical values (e.g., aspect ratios), we express features qualitatively. For example, using terms like ``wider than tall'', ``square-like'', and ``close to the sink'' empirically improved model performance. This strategy aligns with prior findings that language-based tasks are generally more tractable for large language models than compositional or arithmetic reasoning tasks \cite{dziri2023faith,qian2023limitations}.

%This approach builds on prior findings that language models handle language-based tasks more easily than tasks involving composition or arithmetic reasoning.

\paragraph{Prompt Design and Task Framing}

To guide the language model in selecting the most likely storage container for a given item, we explored multiple prompt strategies using ChatGPT-4 \cite{gpt4} and LLaMA-3.3 \cite{llama3}. Each prompt presented a list of container descriptions and required the model to return the most probable container ID, or “None” if none were suitable.

We experimented with three prompt types: (1) Instructional prompts, framing the model as a service robot reasoning about each item; (2) Story-based prompts, encouraging narrative completion grounded in the container descriptions; and (3) Structured system-user prompts, separating global task instructions (system prompt) from instance-specific inputs (user prompt), which yielded the best results. The complete prompt design is provided in Appendix~\ref{sec:NOAM_prompt}.

%===============================================================================

\section{Experimental Setup}
% \begin{itemize}
%     \item \textbf{Methodology:} Dataset split and development.
%     \item \textbf{Baseline Comparisons:} Existing vision-language and object detection models.
%     \item \textbf{Evaluation Metrics:} Accuracy, IoU for object detection.

% \end{itemize}

% \subsection{Methodology: Dataset Split and Development}

We evaluated our method on the real-world dataset (Section~\ref{sec:real}), comparing two model categories: VLMs like Grounding-DINO and MLLMs including Gemini, GPT-4o, Kosmos-2, LLaMA-4, and Qwen-2.5. All models were benchmarked against NOAM.

Model configuration was based on the development set from Section~\ref{sec:dev}, with $5\%$ (370 of 6,500 item-image pairs) used for prompt and parameter tuning. Models were prompted with a household item and tasked with identifying its likely storage location (e.g., drawer, cabinet door) by returning a bounding box. Performance was measured using Intersection over Union (IoU) and accuracy at a binary threshold.

Most experiments ran on a standard Intel Core i7 machine (16GB RAM). GPU-dependent models ran on a university cluster with four NVIDIA GTX 1080$\sim$Ti GPUs (11GB), 125GB RAM, and a 48-core Intel Xeon CPU.

\subsection{Evaluation Metrics}

Each model's performance was evaluated separately on the development and evaluation datasets, and assessed using:

\begin{itemize}
    \item \textbf{Accuracy:} The percentage of predictions with IoU $\geq 0.5$ compared to ground truth.
    \item \textbf{IoU (Intersection over Union):} The overlap between the predicted and true bounding boxes, computed as:
\begin{equation*}
    \text{IoU} = \frac{\text{Area of Overlap}}{\text{Area of Union}}
\end{equation*}
    
\end{itemize}

We set the IoU threshold to 0.5 based on a quality-weighted analysis of human-labeled predictions, where the average IoU of fully or partially correct responses was 0.54. This threshold balances precision and inclusivity, exceeding the upper bound of incorrect predictions and aligning with human judgment. 
%For baselines with prior knowledge of container locations, we set IoU = 1.0. In one case (Grounding-DINO), we also evaluated performance at IoU $\geq 0.95$.
For Grounding-DINO, we additionally evaluated performance at IoU $\geq 0.95$, as its container proposals may slightly differ  warranting a more permissive evaluation alongside the strict 1.0 threshold.
Analysis of the results leading to this value can be found in the Appendix~\ref{sec:iou_threshold}.

%Models relying solely on image-level grounding were evaluated under relaxed IoU conditions ($\geq 0.2$), while those involved in data construction (e.g., Grounding-DINO) were evaluated with stricter thresholds (IoU = 1.0).

\subsection{Baseline Comparisons}
% Full prompts and details per baseline are attached in the Appendix~\ref{sec:baseComp}.

\paragraph{Random} For each item-image pair, we took the list of containers provided by Grounding-DINO during the annotation phase and picked one of them at random.

\paragraph{Grounding-DINO and SAM}

We used Grounding-DINO in combination with SAM to detect and segment storage containers such as drawers and cabinet doors. For this pipeline, we ran the Grounding-DINO detection model with a textual prompt of the form:

\begin{quote}
\texttt{drawer for \{item\} . cabinet door for \{item\}}
\end{quote}

The prompt follows recommendations from the Grounding-DINO authors: it is short, uses simple phrases separated by periods, and avoids complex syntax. The use of ``for'' (e.g., \texttt{drawer for \{item\}}) reflects a concise and structured phrasing style that specifies the relation between an item and a potential storage container.

The following thresholds were applied during inference: \texttt{box\_th-\\reshold = 0.30} and \texttt{text\_threshold = 0.25}, which control the minimum confidence required for detecting bounding boxes and associated text regions, respectively. From the resulting detections, we selected the bounding box with the \textbf{highest confidence} and passed it to the Segment Anything Model (SAM) to generate segmentation masks.

We used the official implementations of both models: \path{IDEA-Research/grounding-dino} and \path{facebook/sam-vit-h}. Grounding-DINO was used at commit \texttt{c023468} of the \texttt{IDEA-Research/grou-\\nding-dino} repository. The SAM implementation corresponds to the official April 2023 release from Meta\footnote{\url{https://github.com/facebookresearch/segment-anything}}, using the \texttt{vit-h} variant.

Grounding-DINO handled object-level bounding box detection conditioned on the textual prompt, while SAM produced binary segmentation masks for those regions. The segmentation was implemented using the SAM PyTorch interface, which transforms bounding boxes to input-specific image coordinates and outputs a binary mask for each box.

The segmentation masks were post-processed using OpenCV to extract polygon contours. We applied the Ramer-Douglas-Peucker (RDP) algorithm with \texttt{epsilon = 0.02 * arcLength} to simplify the polygons.

It is important to note that this use of Grounding-DINO and SAM is distinct from the initial use of Grounding-DINO during dataset annotation, where all visible containers (drawers and cabinets) were detected in bulk. In the current evaluation phase, we specifically query for each item separately, prompting the model to localize where the item is likely stored. To ensure a fair evaluation without introducing annotation bias, a prediction was only considered correct if it achieved an IoU $\geq 0.95$ to allow only small correction with the ground-truth container.

\paragraph{Grounding‑DINO and SAM without Item Prompt}  
To assess the influence of including the item name, we also ran Grounding‑DINO with a minimal prompt - identical to the annotation phase - omitting the item entirely:

\begin{quote}
\texttt{drawer . cabinet door}
\end{quote}

Again, we selected the container with the highest confidence score. Comparing these results to those obtained with the item‑inclusive prompt allows us to quantify how much the item name boosts the model’s performance.

\paragraph{Kosmos-2}

We used the Kosmos-2 MLLM with both image and text input. The prompt followed a structured grounding format using the special tokens provided by the model:

\begin{quote}
\texttt{<grounding> In which<phrase> drawer</phrase> or<phrase> cabinet door</phrase> is<phrase> a \{item\}</phrase> stored?}
\end{quote}

The prompt structure follows Kosmos-2’s grounding format, where the \texttt{<grounding>} tag activates grounding mode and each \texttt{<phrase>} tag marks a semantically important span. By marking both the container types (e.g., \texttt{drawer}, \texttt{cabinet door}) and the queried item, the model is guided to focus on relevant visual concepts and return grounded predictions accordingly.

The model's outputs captions and entity bounding boxes relevant to the query. Since Kosmos-2 does not use a pre-defined list of container locations (unlike human annotators), we considered a prediction correct if the IoU between the model's output and the ground truth was $\geq 0.5$ (meaning that the bounding box overlapped the true container in at least $20\%$ of its area).

We used the official HuggingFace implementation\footnote{\url{https://huggingface.co/microsoft/kosmos-2-patch14-224}}. For inference, we used the model's \texttt{generate()} method with the following settings:

\begin{itemize}
    \item \texttt{use\_cache=True}: enables reuse of key and value tensors from previous self-attention layers for efficient autoregressive decoding.
    \item \texttt{max\_new\_tokens=128}: limits the number of newly generated tokens to 128.
    \item \texttt{skip\_special\_tokens=True}: ensures that decoding omits any special tokens in the output.
\end{itemize}

After generation, we applied the \texttt{post\_process\_generation()} method twice: first with \texttt{cleanup\_and\_extract=False} to inspect raw model output, and again with default settings to extract grounded entity descriptions and their associated bounding boxes.

All inference was conducted on a single NVIDIA GPU, using PyTorch. We used the pre-trained checkpoint without any additional fine-tuning or modification.

\paragraph{Gemini (Google MLLM)}

We used Google’s Gemini API to evaluate multimodal performance. Specifically, we tested two model versions: 
\texttt{gemini-\allowbreak 1.5-flash-latest} and \texttt{gemini-2.5-\allowbreak flash}. Predictions were considered correct if the IoU with ground truth was $\geq 0.5$, as in the Kosmos-2 evaluation.

Each image-item pair was processed using the following template:

\begin{quote}
\texttt{Analyze the provided image of a kitchen.\\
Identify the item: '\{item\}'\\
Determine the most likely storage location for this item, considering only drawers or cabinet doors visible in the image.\\
Provide the bounding box coordinates for this storage loc-\\ation as a Python list of four integers: [x\_min, y\_min, x\_max, y\_max].\\
The coordinates should be relative to the image dimensions (top-left is [0,0]).\\
Only output the list of coordinates and nothing else. For example: [100, 200, 300, 400]\\
If you cannot determine a likely storage location (drawer or cabinet door) for this item, output: [0, 0, 0, 0]}
\end{quote}

The API was accessed via an official Google API key, with reliability safeguards including up to 3 retry attempts (starting with a 5-second delay), and a 1-second pause between successful calls. We saved progress every 50 processed examples.

\paragraph{ChatGPT-4o}

We used the GPT-4o model with vision capabilities through the OpenAI API. Predictions were considered correct if the IoU with the ground-truth polygon was $\geq 0.5$, consistent with other models.

The prompt format was:

\begin{quote}
\texttt{You are analyzing a kitchen image.\\
The visible containers are drawers and cabinet doors.\\
The item to store is: \{item\}\\
Determine the most likely storage location among visible drawers and cabinet doors only.\\
Return a list of 4-point bounding box coordinates for the item.\\
If a suitable location cannot be determined, return an empty list [].\\
Only return the bounding box list, nothing else.}
\end{quote}

The API was accessed via an OpenAI API key. For each image-item pair, the image was encoded in base64 and sent alongside the prompt in a multimodal request. Responses were parsed to extract the bounding box list.
Robustness measures included handling exceptions per request and adding a 1-second delay between API calls to avoid rate limits.

\paragraph{LLaMA-4 and Qwen-2.5 (Together AI API)}

We evaluated the multimodal LLaMA-4 and Qwen-2.5 models via the Together AI API using the same prompt and base64 image format as GPT-4o. Specifically, we employed the checkpoints \texttt{meta-llama/Llama-4-Maverick-17\\B-128E-Instruct-FP8} and \texttt{Qwen/Qwen2.5-VL-72B-Instruct}, respectively. As with the other models, predictions were considered correct if the IoU with the ground-truth container polygon was $\geq 0.5$.
% 
% The prompt format was identical to our GPT‑4o setup:
% 
% \begin{quote}
% \texttt{You are analyzing a kitchen image.\\
% The visible containers are drawers and cabinet doors.\\
% The item to store is: \{item\}\\
% Determine the most likely storage location among visible drawers and cabinet doors only.\\
% Return a list of 4-point bounding box coordinates for the item.\\
% If a suitable location cannot be determined, return an empty list [].\\
% Only return the bounding box list, nothing else.}
% \end{quote}

Each request included the Base64‑encoded image and the text prompt in a single multimodal payload. We accessed the API using a Together AI API key, parsed the JSON response to extract the bounding box list, and converted it to our evaluation format.

%===============================================================================

\section{Results and Analysis}
\label{sec:result}
% \begin{itemize}
%     \item Performance of both models on stored object retrieval.
%     \item Ablation studies on different object categories and storage types.
%     \item Generalization capability across different households.
% \end{itemize}

% \paragraph{Performance on stored object retrieval}

\paragraph{Real-World Evaluation Dataset} 
We begin by reporting the overall performance of the various algorithms on the real-world evaluation dataset. As shown in Table~\ref{tab:test_results}, NOAM achieves the highest accuracy and IoU among all of the evaluated models, outperforming random, vision-based, and multimodal baselines. 

\begin{table}[t]
\centering
\caption{Evaluation Set Accuracy and Average IoU for Various Models.}
\label{tab:test_results}
\footnotesize
\begin{tabular}{lcc}
\toprule
\textbf{Model} & \textbf{Accuracy (\%)} & \textbf{Average IoU} \\
\midrule
Human Annotator 1 (IoU = 1) & 38.00 & 0.380 \\
Human Annotator 2 (IoU = 1) & 27.00 & 0.271 \\
Human Annotator 3 (IoU = 1) & 36.00 & 0.361 \\ \\
Random (IoU = 1)   & 6.00  & 0.062 \\
Grounding-DINO (IoU = 1) & 13.00 & 0.188 \\
Grounding-DINO (IoU $\geq$ 0.95) & 17 & 0.188 \\
Grounding-DINO - \\ no item in prompt (IoU = 1) & 10 & 0.117 \\
Kosmos-2 (IoU $\geq$ 0.5)              & 4.00  & 0.042 \\
Gemini-1.5-flash (IoU $\geq$ 0.5)                & 3.00  & 0.034 \\
Gemini-2.5-flash (IoU $\geq$ 0.5)                & 1.00  & 0.027 \\
GPT-4o API (IoU $\geq$ 0.5)            & 8.00  & 0.082 \\
LLaMA-4 (IoU $\geq$ 0.5)            & 1.00  & 0.094 \\
Qwen-2.5 (IoU $\geq$ 0.5)            & 5.00  & 0.091 \\
\textbf{NOAM LLaMA-3.3 (IoU = 1)}     & \textbf{23.00} & \textbf{0.232} \\
\textbf{NOAM GPT-4 (IoU = 1)}     & \textbf{23.00} & \textbf{0.23} \\
\bottomrule
\end{tabular}
\end{table}

For follow-up statistical tests, we conducted a one-way ANOVA on accuracy scores, followed by post-hoc pairwise comparisons using the Bonferroni correction. The results confirmed a significant effect of model type on performance ($p < 0.05$). Specifically, NOAM GPT‑4 significantly outperformed Gemini‑1.5‑flash, Gemini‑2.5‑flash, and Kosmos‑2 after Bonferroni correction. Differences with other models, including GPT‑4o, Grounding‑DINO, LLaMA‑3.3, and LLaMA‑4, were not statistically significant. NOAM also did not differ significantly from human annotators.

While NOAM did not reach human-level performance, its accuracy ($23\%$) is notably closer to that of the least accurate human annotator ($27\%$) than the gap between that annotator and the next-best human ($36\%$). Further, the pairwise comparison between NOAM and the least accurate human annotator did not reach statistical significance ($p = 0.006, 0.46, 0.026$ for annotators 1, 2, and 3 respectively, before the Bonferroni correction and $p = 0.67, 1.0, 1.0$ after the correction). This result is encouraging and suggests that the pipeline of NOAM has strong potential to soon reach human-level performance, especially as new VLMs and MLLMs become available for integration.

% \paragraph{Item-Level Performance Variability} 
% We analyzed per-item model performance to uncover finer-grained differences. Sample sizes ranged from 1 to 44, see Appendix~\ref{sec:item_preformance}). A repeated-measures ANOVA on per-item found no significant accuracy differences between models or human annotators, suggesting comparable overall performance.
% To assess model consistency across items, we visualized per-item accuracy using a clustered column chart (Figure~\ref{fig:per_item_acc}). Human annotators outperformed all models on Tupperware containers, while NOAM variants were the most consistent across items, especially on bottle openers and painkillers.

% \begin{figure}[t]
%     \centering
%     \includegraphics[width=1\linewidth]{AAAI/per_item_accuracy.png}
%     \caption{Clustered column chart showing per-item accuracy (\%) for each baseline across four representative items.}
%     \label{fig:per_item_acc}
% \end{figure}

% To assess model consistency across items, we counted how often each model ranked among the top-3 performers per item (see Figure~\ref{fig:top_3_model_counts}). Human annotators led with five top-3 appearances, followed by NOAM GPT-4 and NOAM LLaMA-3.3, each with four appearances.

% \begin{figure}[t]
%     \centering
%     \includegraphics[width=1\linewidth]{top_3_model_counts.png}
%     \caption{Number of times each model ranked among the top-3 per item (accuracy $>$ 0 only).}
%     \label{fig:top_3_model_counts}
% \end{figure}

\paragraph{Development Dataset} 
Table~\ref{tab:dev_results} presents results on the annotated development set. While our focus is on evaluation performance, development scores help reveal whether models learned meaningful patterns. For instance, Kosmos-2 underperforms even on the development set (below random) suggesting limited capacity or misalignment with the task rather than overfitting. In contrast, NOAM performs strongly, reflecting effective task-specific learning. Human scores are based on overlapping annotations used for agreement analysis and do not reflect full dataset performance.

% We continue by reporting the accuracy and IoU results on the annotated development dataset, as shown in Table \ref{tab:train_results}. While our primary focus is on evaluation set results, the development set performance helps reveal whether a model has learned meaningful patterns or is simply underfitting. For example, models like Kosmos-2 and Gemini achieve relatively low accuracy even on the development set (lower than random from a given set of containers), suggesting limited capacity or misalignment with the task rather than overfitting.  Including development results thus helps distinguish between models that fail to generalize and those that fail to learn effectively in the first place.
% Unsurprisingly, human annotators achieve very high training performance, as the training data is constructed based on their annotations (as discussed in Section \ref{sec:real}). In this dataset, NOAM also achieves substantially higher training performance than the other models.
%, highlighting the benefits of task-specific reasoning and structured input.

\begin{table}[t]
\centering
\footnotesize
\caption{Development Set Accuracy and Average IoU for Various Models.}
\label{tab:dev_results}
\begin{tabular}{lcc}
\toprule
\textbf{Model} & \textbf{Accuracy (\%)} & \textbf{Average IoU} \\
\midrule
%Human Annotator 1 (IoU = 1) & 87.52 & 0.876 \\
%Human Annotator 2 (IoU = 1) & 85.63 & 0.857 \\
%Human Annotator 3 (IoU = 1) & 86.33 & 0.863 \\
Human Annotator (Estimated) & 35.83 & 0.361 \\
Random (IoU = 1)   & 7.55  & 0.083 \\
Grounding-DINO (IoU = 1) & 5.89 & 0.112 \\
Grounding-DINO (IoU $\geq$ 0.95) & 9.49 & 0.112 \\
Grounding-DINO - \\ no item in prompt (IoU = 1) & 11.35 & 0.128 \\
Kosmos-2 (IoU $\geq$ 0.5)              & 0.46  & 0.013 \\
Gemini-1.5-flash (IoU $\geq$ 0.5)                & 8.11  & 0.083 \\
Gemini-2.5-flash (IoU $\geq$ 0.5)                & 2.43  & 0.03 \\
GPT-4o API (IoU $\geq$ 0.5)    & 13.78  & 0.138 \\
LLaMA-4 (IoU $\geq$ 0.5)    & 3.24  & 0.071 \\
Qwen-2.5 (IoU $\geq$ 0.5)    & 7.57  & 0.099 \\
\textbf{NOAM LLaMA-3.3 (IoU = 1)}        & \textbf{23.51} & \textbf{0.244} \\
\textbf{NOAM GPT-4 (IoU = 1)}        & \textbf{28.11} & \textbf{0.287} \\
\bottomrule
\end{tabular}
\end{table}

\section{Discussion and Future Work}
NOAM demonstrates strong reasoning abilities, but several aspects warrant further development:

\noindent\textbf{Efficiency and Scalability}
NOAM’s average inference time is approximately 13 seconds per image. The main bottlenecks are Grounding-DINO, which accounts for about 10 seconds, and the LLM API call (e.g., ChatGPT), which takes around 2.9 seconds. Using the LLaMA-3.3 (via Together-AI API) slightly increases total inference time to roughly 15.9 seconds per image. %These timings include network overhead and client-side processing (the net server-side processing time is not exposed by providers).

While this runtime is acceptable for research and offline applications, real-world deployment would require further optimization. Initial experiments replacing Grounding-DINO with a lightweight detector reduced detection time to under 1 second. Future directions for improving efficiency include model distillation, prompt caching, and the use of smaller, fine-tuned modules to support near real-time inference.
%NOAM’s $\sim$13 seconds per image inference is bottlenecked by Grounding-DINO and LLM APIs. Replacing Grounding-DINO with a lightweight detector cut detection time to $\sim$1s. Future improvements include model distillation, caching, and smaller, fine-tuned modules to support real-time use.

\noindent\textbf{Generalization and Dataset Diversity}
Performance is expected to vary by image domain. Expanding benchmarks to other room types (e.g., bedrooms, industrial scenes) and adapting anchors and prompts will test robustness.

\noindent\textbf{Embodied and Interactive Reasoning}
NOAM currently uses static images. Extending it to multi-view or interactive setups could reduce occlusion errors and support belief updates during exploration.

\noindent\textbf{Model Robustness}
LLM outputs vary by prompt and model. We plan to fine-tune VLM or MLLM on our dataset to improve consistency and compare against zero-shot performance.

\noindent\textbf{Richer Priors and Personalization}
Integrating spatial priors (e.g., furniture hierarchies) and user-specific data (e.g., gestures or preferences) may enhance practical accuracy and personalization.

\noindent\textbf{Practical Metrics}
IoU alone is insufficient for real-world robotics. Future work will incorporate task-based metrics (e.g., success rate, energy use), and propose container selection heuristics to improve robustness to detection errors.

%===============================================================================
\section{Conclusion}
% \begin{itemize}
%     \item Summary of contributions.
%     \item Potential applications in real-world service robotics.
%     \item Future directions: Enhancing robustness, incorporating user preferences.
% \end{itemize}

This paper introduces the Stored Household Item Challenge, a novel problem designed to evaluate semantic spatial reasoning about the likely locations of non-visible objects in household environments. Unlike conventional object detection tasks, this challenge targets a critical but underexplored capability: inferring hidden item locations based on context and commonsense knowledge. We present two new datasets of labeled item-image pairs featuring concealed storage scenarios, along with a suite of baseline evaluations. 
To tackle this task, we further propose NOAM (Non-visible Object Allocation Model), a vision-to-language pipeline that reformulates storage prediction as a structured natural language inference problem. Experimental results demonstrate that NOAM significantly outperforms vision-only baselines, though still not comparable to human-level reasoning capabilities.

This work opens several directions for future research. For example, NOAM currently relies on static descriptions and a single scene view. Future models could benefit from richer inputs, such as sequential observations, user preferences, or multimodal feedback. Expanding the spatial reasoning framework to include hierarchical priors or soft constraints (e.g., ``items are often above counters'') could further improve performance. Additionally, integrating this task into interactive, embodied agents may allow systems to refine predictions through exploration and reinforcements, enhancing robustness in real-world settings.

Although this paper focuses on kitchens, the underlying reasoning principles generalize to a wide range of domestic and industrial environments. We plan to expand the dataset and task to include additional spaces and a broader set of item categories. By advancing models of hidden object reasoning, this work contributes to building service robots that can function effectively in unfamiliar, dynamic, and complex human environments.

\bibliographystyle{ACM-Reference-Format} 
\bibliography{bibliography}

\clearpage 
\appendix
\section{NOAM: Prompt Design and Task Framing}
\label{sec:NOAM_prompt}
After generating descriptions for all containers in an image, we focused on designing an effective prompt to convey the task to the language model. All prompt variants were tested using ChatGPT-4, and the final selected prompt was also applied to LLaMA-3.3.

In every prompt format and for both models, the goal was to select the most likely container ID from the provided list, indicating where the queried item would typically be stored. If more than one container was retrieved (in case that two containers were selected as the most likely), only the first was used in the evaluation. If the model determined that none of the containers were suitable, it returned ``None''.

Below are the main prompt strategies we explored:

%\vspace{0.5em}
\subsubsection*{1. Instructional Prompt: Service Robot Role}
Our first approach described the model as a service robot trying to locate an item in a kitchen image, based on a list of container descriptions:

\begin{lstlisting}
prompt = f"""You are a service robot in a domestic environment. You are now looking at a kitchen scene.

The containers (drawers, cabinet doors etc.) detected in the image are:
{chr(10).join(f"- {desc}" for desc in containers)}

For each of the following items:
{', '.join(items)}

Please determine in which container each item is most likely to be stored. If no container is suitable, say so. Provide reasoning for each item.

Format:
Item: [Name]
Best container: [just container id or "None"]
Reasoning: [Your explanation]
"""
\end{lstlisting}

%\vspace{0.5em}
\subsubsection*{2. Story-Based Prompt}
We next tried a narrative approach, asking the model to complete a short story that ends with a storage decision:

\begin{lstlisting}
prompt = (
   f"As they stepped into the kitchen, they began searching for a {item.lower()}, scanning the scene for storage areas - known as containers, like drawers, cabinets, or pantry doors where household items are typically kept.\n"
   "Based on the descriptions of the detected containers, they paused in front of one that seemed just right and reached toward...\n"
   "\n"
   "Finish the story by selecting the most suitable container from the list below, or say the item isn't in any container if none of them are appropriate:\n"
   f"{chr(10).join(f"- {desc}" for desc in containers)}"
)
\end{lstlisting}

Although this version was more natural and creative, it generally underperformed compared to the direct instructional format.

%\vspace{0.5em}
\subsubsection*{3. Structured Prompt with System and User Roles}
After learning more about prompt engineering, we adopted a structured format to balance informativeness and brevity. This prompt format allowed the model to internalize the task logic from the system message and focus on inference during deployment. It consists of two parts:
\begin{itemize}
    \item \textbf{System Prompt:} Encodes general knowledge about the task, including comprehensive instructions and illustrative examples to prime the model.
    \item \textbf{User Prompt:} Provides a concise task instance, including the item to be located and the textual descriptions of all detected containers in the scene as presented above.
\end{itemize}

\begin{lstlisting}
system_prompt = """
You are helping locate a household item in a kitchen.
The item is stored in one of several visible containers (e.g., drawers, cabinets), but I don't know which.
I'll provide a list of container descriptions and the item name.
Your task is to identify the most likely container based on typical kitchen organization. If none are suitable, return "None".

Response format:
Item: [Name]  
Best container: [Container ID or "None"]  
Reasoning: [Short explanation]

### Example 1
Item: Fork  
Containers:  
- Container 1: cabinet door below the countertop, located to the right of the dishwasher.  
- Container 2: below the countertop, located to the left of the dishwasher.  
- Container 3: cabinet door above the countertop, located above the coffee machine.  
- Container 4: drawer below the countertop.  
- Container 5: cabinet door above the countertop.  
- Container 6: cabinet door above the countertop.  

Item: Fork  
Best container: 4  
Reasoning: Forks are usually stored in drawers below the countertop for easy access.

### Example 2
Item: Trash Bag  
Containers:  
- Container 1: cabinet door above the countertop.  
- Container 2: drawer below the countertop.  
- Container 3: cabinet door below the countertop, located below the sink.  

Item: Trash Bag  
Best container: 3  
Reasoning: Trash bags are commonly stored under the sink near the trash can.

### Example 3
Item: Knife  
Containers:  
- Container 1: cabinet door.  
- Container 2: drawer, located below the electronic kettle, above the oven, at the bottom-right of the refrigerator, and at the bottom-left of the dish drying rack.  
- Container 3: cabinet door, located below the dishwasher.  

Item: Knife  
Best container: 2  
Reasoning: Knives are typically stored in drawers for safety and accessibility.

### Example 4
Item: Baking pan  
Containers:  
- Container 1: drawer below the countertop, located at the bottom-right of the oven, and at the top-right of the electronic kettle.  
- Container 2: cabinet door below the countertop, located to the right of the electronic kettle, and at the bottom-right of the oven.  
- Container 3: cabinet door below the countertop, located below the stove.  
- Container 4: below the countertop.  
- Container 5: cabinet door below the countertop, located to the left of the electronic kettle, and at the bottom-left of the oven.  

Item: Baking pan  
Best container: 5  
Reasoning: Baking pans are stored in cabinets below the countertop near the oven.

Now respond to the following:
""".strip()

user_prompt = f"Item: {item}\nContainers:\n" + "\n".join(f"- {desc}" for desc in containers)
\end{lstlisting}

\section{Experimental Setup: Determining IoU threshold}
\label{sec:iou_threshold}
To select an appropriate IoU threshold for our results, we first divided the models and baselines into two groups:

\begin{itemize}
    \item \textbf{Models with prior container knowledge:} This includes human annotators (who chose from a predefined list of containers in the web application), the random baseline (which selects a random container from Grounding-DINO detections), Grounding-DINO with the item in the prompt (used both for object detection and for item storage), and NOAM (which also receives the list of detected containers).
    \item \textbf{Models without prior container knowledge:} This includes Gemini, GPT-4o, Kosmos-2, LLaMA-4, and Qwen-2.5.
\end{itemize}

For fair comparison, we required IoU=1.0 for the first group. For the second group, we examined cases with IoU $\geq 0.0$ and IoU $\geq 0.1$, performing a sanity check by reviewing all images, marking the ``correct'' container polygons, and assigning scores of 0 (incorrect), 0.5 (partially correct), or 1 (fully correct). Computing a quality‑weighted average IoU from these scores yielded 0.54, which guided our choice of 0.5 as the threshold.

\begin{figure}[H]
    \centering
    \includegraphics[width=1\linewidth]{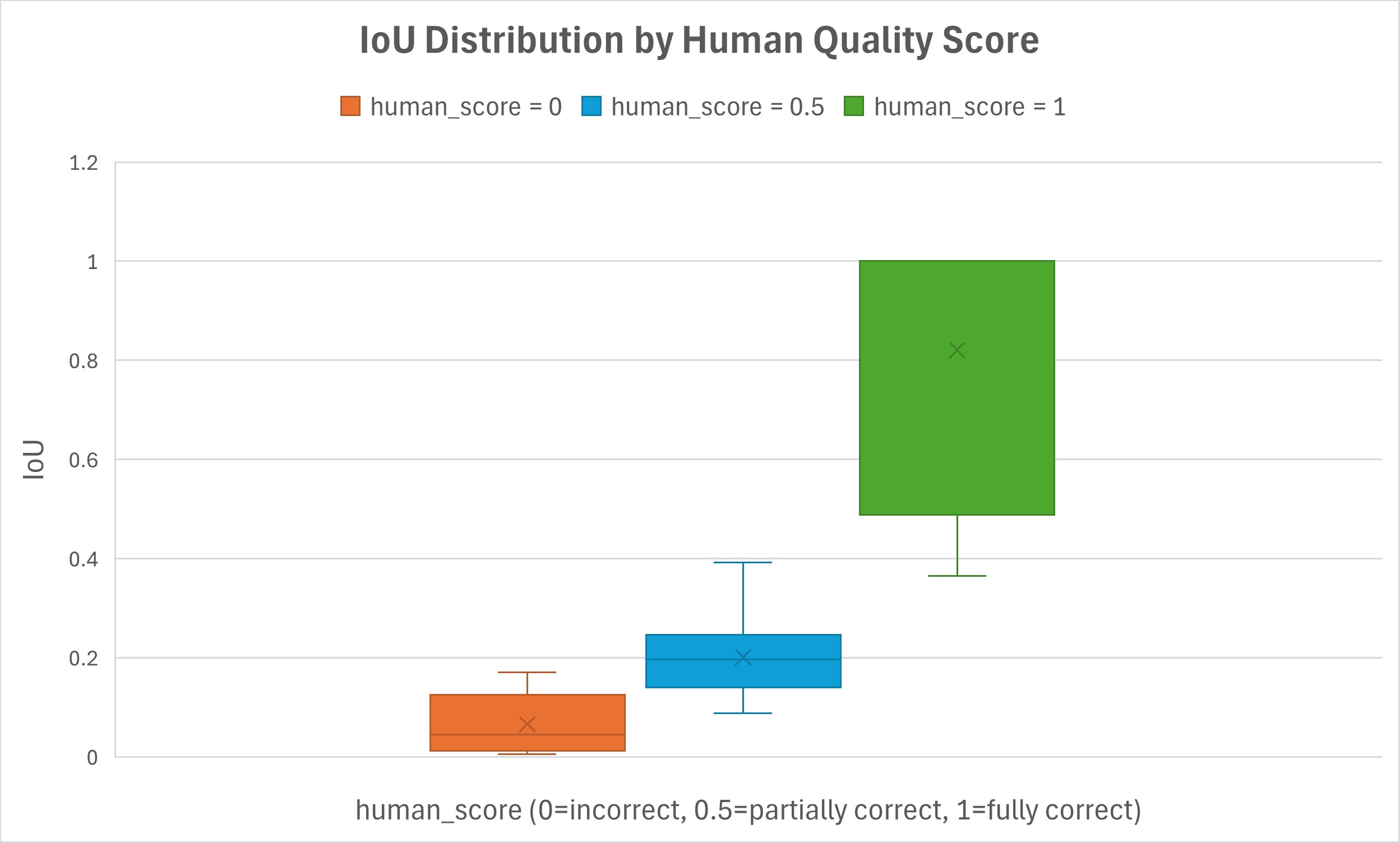}
    \CaptionWithDescription{IoU distributions across human-assigned quality scores (0=incorrect, 0.5=partially correct, 1=fully correct).}
    \label{fig:IoU_dist}
\end{figure}

As shown in Figure~\ref{fig:IoU_dist}, incorrect predictions (score = 0) never exceed IoU=0.17, while fully correct ones (score = 1) never fall below IoU $\approx 0.365$. Partially correct cases lie between IoU $\approx 0.088$ and 0.392. This gap suggests an ideal cutoff in [0.17, 0.365]. We adopt IoU $\geq 0.5$, which is stricter than the minimum correct bound, to maximize precision based on our quality‑weighted analysis.

\end{document}